\documentclass[10pt,twocolumn,letterpaper]{article}

\usepackage{cvpr}              % To produce the CAMERA-READY version
\usepackage{graphicx}
\usepackage{amsmath}
\usepackage{amssymb}
\usepackage{booktabs}
\usepackage{multirow}
\usepackage{booktabs}
\usepackage{pifont}
\usepackage{bbding}
\usepackage[numbers,sort]{natbib} 
\usepackage{color}

\usepackage[pagebackref,breaklinks,colorlinks]{hyperref}

\usepackage[capitalize]{cleveref}
\crefname{section}{Sec.}{Secs.}
\Crefname{section}{Section}{Sections}
\Crefname{table}{Table}{Tables}
\crefname{table}{Tab.}{Tabs.}

\def\cvprPaperID{4573} % *** Enter the CVPR Paper ID here
\def\confName{CVPR}
\def\confYear{2022}

\begin{document}

\title{Modeling Motion with Multi-Modal Features for Text-Based Video Segmentation}

\author{Wangbo Zhao$^{1,2,3}$\quad 
Kai Wang$^{1}$\quad
Xiangxiang Chu$^{2}$\quad
Fuzhao Xue$^{1}$\quad
Xinchao Wang$^{1}$\quad
Yang You$^{1}$\footnotemark[1]\quad\\
$^1$ National University of Singapore \quad
$^2$ Meituan Inc. \quad
$^3$ Northwestern Polytechnical University \\
 {\tt\small
wangbo.zhao96@gmail.com, kai.wang@comp.nus.edu.sg, chuxiangxiang@meituan.com,}\\
{\tt\small f.xue@u.nus.edu, xinchao@nus.edu.sg, youy@comp.nus.edu.sg}}

\maketitle
\footnotetext[1]{Corresponding author.}

\vspace{-0.1cm}
\begin{abstract} 
Text-based video segmentation aims to segment the target object in a video based on a describing sentence. Incorporating motion information from optical flow maps with appearance and linguistic modalities is crucial yet has been largely ignored by previous work. In this paper, we design a method to fuse and align appearance, motion, and linguistic features to achieve accurate segmentation. Specifically, we propose a multi-modal video transformer, which can fuse and aggregate multi-modal and temporal features between frames. Furthermore, we design a language-guided feature fusion module to progressively fuse appearance and motion features in each feature level with guidance from linguistic features. Finally, a multi-modal alignment loss is proposed to alleviate the semantic gap between features from different modalities. Extensive experiments on A2D Sentences and J-HMDB Sentences verify the performance and the generalization ability of our method compared to the state-of-the-art methods.
\end{abstract}

\footnotetext{Our code is publicly available at: \url{https://github.com/wangbo-zhao/2022CVPR-MMMMTBVS}.}

\vspace{-6mm}  
\section{Introduction} \label{Introduction}
\vspace{-3mm}
Text-based video segmentation aims at locating and segmenting the object described by a language sentence in a video sequence. Unlike traditional tasks, which do prediction on video- or frame- level, \eg text-to-video retrieval \cite{xu2016msr, krishna2017dense, rohrbach2015dataset}, video caption \cite{zhou2018towards, li2016tgif}, video question answering \cite{xu2017video, jang2017tgif}, and language-queried video localization \cite{anne2017localizing,zhang2021natural}, this task requires relatively more fine-grained multi-modal and temporal understanding for pixel-level segmentation. The challenge of this task can be thus summarized as: (1) how to reason between visual and linguistic modalities to locate the target object, and (2) how to leverage temporal information to enhance segmentation.

\begin{figure}[!t]
    \centering
    \includegraphics[width=1\linewidth]{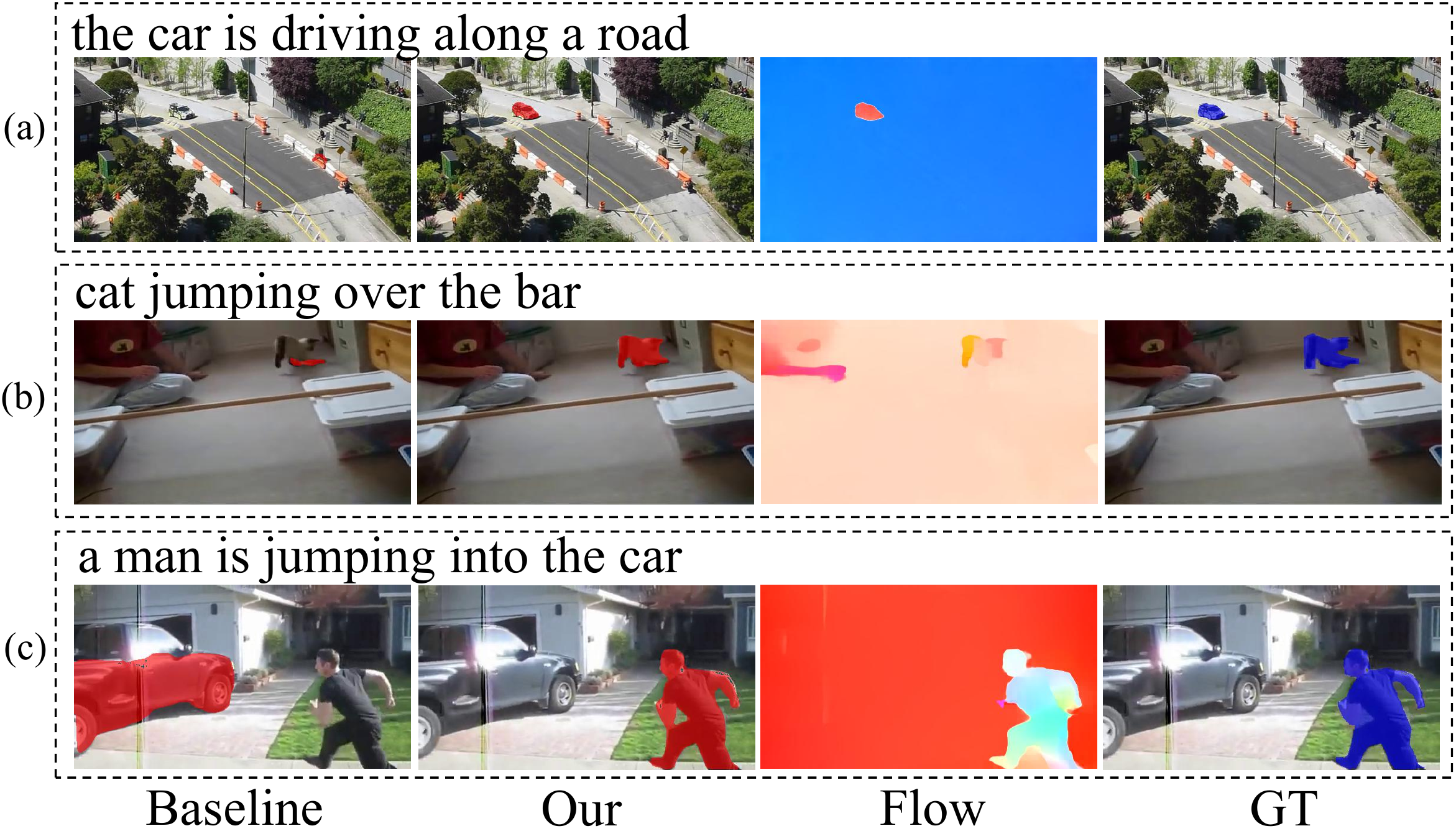}
    \caption{Comparison between baseline and our model. We adopt "B" in \ref{albation study} as the baseline model. Compared with the baseline model, our model can incorporate motion information from optical flow maps with appearance and linguistic features and generate better segmentation masks.}
    \label{compare}
    \vspace{-7mm}
\end{figure}

To solve the former problem, previous works adopt simple concatenation \cite{hu2016segmentation}, generating dynamic filters \cite{gavrilyuk2018actor, wang2020context} and cross-modal attention modules \cite{hui2021collaborative, wang2019asymmetric} to achieve interactions between two modalities. When it comes to the latter problem, they usually adopt 3D convolution neural networks (3D CNNs) \eg I3D \cite{carreira2017quo} to extract features from a video clip. However, all these methods ignore exploring the explicit motion information between frames for text-based video segmentation. In this task, the target object usually has action, and the corresponding text contains some words to describe its motion \eg driving and jumping in Figure~\ref{compare}. This means that the motion information may help the model to find the target object. Despite the fact that some motion information between frames can be implicitly learned in 3D CNNs, it can not well interact with other modalities. Introducing motion information has been tried in some video tasks \cite{ji2021full, zhao2021weakly, chen2021mm, li2019motion, dutt2017fusionseg, zhou2020motion, yang2021self}, but how to incorporate the motion information with appearance and linguistic features in text-based video segmentation is still challenging.

A common way to introduce explicit motion information is to extract features from flow maps generated from an optical flow estimation model. From flow maps in Figure~\ref{compare}, we can find that the target object with motion usually is distinctive and can be easily identified. This may promote the final performance. To leverage the motion information from optical flow, Gavrilyuk \etal \cite{gavrilyuk2018actor} adopt two 3D CNNS with different parameters to generate masks from RGB frames and optical flow maps, respectively, then compute weighted averaged masks from them. However, such a simple fusion strategy ignores the interaction between motion modalities and appearance and linguistic features, leading to unsatisfactory improvement and huge computational overhead. Hence, designing a model to effectively incorporate the motion information from the optical flow with appearance features from RGB frames and linguistic features is necessary.

Motivated by observations above, we propose our multi-modal fusion and alignment network. First, since many previous works \cite{chen2021mm, hu2021unit, bertasius2021space, arnab2021vivit} have demonstrated the superiority of transformers in reasoning and fusing multi-modal and temporal features, we build a multi-modal video transformer (MMVT) to model the interaction between appearance, motion, and linguistic features in different frames. Our transformer contains two attention modules in each layer: cross-modal attention and temporal attention. The former aims at fusing three modalities features, while the latter is adopted to aggregate fused features in the temporal dimension. By stacking them several layers, multi-modal information can flow and interact with each other between different frames. Benefiting from the multi-modal interaction between frames in MMVT, we do not rely on 3D CNNs to extract temporal information, which largely reduce the computational overhead.

Then, to fuse multi-modal features progressively, we propose the language-guided feature fusion (LGFF) module and insert it into each level to decode features. In each module, useful appearance and motion features will be selected by the linguistic feature, with the help of features from the higher level. By doing this, useful features can be gradually selected and fused. Moreover, since both appearance, motion, and language features are distinctive modalities features, which are generated from backbones separately pre-trained on different datasets, the semantic gap between them would be large \cite{huang2020pixel}. To alleviate this problem, we design a multi-modal alignment loss, which explicitly encourages the network to learn to align three modalities features in an embedding space, which further improves the performance of our model.

In Figure~\ref{compare}, compared with the baseline model without motion information, our model can accurately locate the target object, obtain a more complete mask, and distinguish the target object from others. Our main contributions can be summarised as:
   \vspace{-2mm}
\begin{itemize}
   \item To the best of our knowledge, we are the first to incorporate the motion information from optical flow maps with appearance and linguistic features for text-based video segmentation.
\vspace{-2mm}
    \item We propose a transformer-based model to fuse multi-modal and temporal features and design a language-guided feature fusion module to progressively fuse multi-modal features from different feature levels.
\vspace{-2mm}
    \item Noticing the semantic gap between different modal features, we propose a multi-modal alignment loss to explicitly align features from three different modalities, which further improve the performance of our method.
\vspace{-2mm}    
    \item Extensive experiments are conducted to verify the effectiveness of proposed methods. Our approach significantly surpasses existing state-of-the-art methods on most metrics on A2D Sentences and J-HMDB Sentences dataset with less computational overhead.
\end{itemize}

\vspace{-4mm}  
\section{Related Work}
\vspace{-2mm}
\noindent\textbf{Text-Based Image Segmentation} % ️✔️
Text-based image segmentation aims to segment the object in an image given a text describing its properties \eg appearance and location.  Hu \etal \cite{hu2016segmentation} are the first to propose this task, and they adopt the fully convolutional network to fuse extracted visual and linguistic features directly. Liu \etal \cite{liu2017recurrent} propose a multi-modal LSTM to force the word-visual interaction. Ye \etal \cite{ye2019cross} design a self-attention module to capture long-range relationships between two modalities. Luo \etal \cite{luo2020multi} propose a model to achieve joint learning of locating and segmentation since these two tasks can reinforce each other. Jing \etal \cite{jing2021locate} decouple this task into locating the target object position and accurately generating the segmentation mask. Yang \etal \cite{yang2021bottom} represents the expression as a language graph and performs explainable visual reasoning to distinguish the target object from others. Ding \etal \cite{ding2021vision} introduce the encoder-decoder attention mechanism in transformer \cite{vaswani2017attention} and view the language expression as queries.

Unlike these works for images, which only need to focus on fusing features from the static RGB image and the language expression, we conduct multi-modal fusion between the RGB image, flow map, and text. In addition, we also consider the temporal information between adjacent frames.

\noindent\textbf{Text-Based Video Segmentation} % ️️✔️
For promoting comprehensive action understanding, Xu \etal \cite{xu2015can} release a dataset named Actor-Action Dataset (A2D) containing a fixed vocabulary of actor and action pairs and pixel-level annotations. After that, Gavrilyuk \etal \cite{gavrilyuk2018actor} further extend this dataset and propose text-based video segmentation. They generate dynamic filters from extracted text features and adopt them to convolve with vision features to obtain the final pixel-wise segmentation.
They also try to average the masks from an optical flow map and an RGB frame to improve the performance further. Wang \etal \cite{wang2019asymmetric} propose a cross-guided attention mechanism, where features from frames and the text can guide and promote each other. This design can reduce linguistic variation and incorporate query-focused visual features. Mcintosh \etal \cite{mcintosh2020visual} propose a capsule-based network to encode and merge visual and textual features jointly. Wang \etal \cite{wang2020context} introduce the idea of deformable convolution \cite{dai2017deformable} into generating dynamic filters to address geometric deformation. Ning \etal \cite{ning2020polar} propose a polar positional encoding mechanism to measure the spatial relations in terms of direction and range, which is similar to natural language descriptions. Hui \etal \cite{hui2021collaborative} adopt 3D and 2D encoders to recognize the queried actions and accurately segment target object, respectively.

Different from \cite{gavrilyuk2018actor}, which ignores the interaction between motion information and other modalities, the motion information can be well fused and interact with appearance and linguistic features in our MMVT and LGFF.

\noindent\textbf{Vision-Language Learning Tasks} % ️️✔️
Owing to the development of NLP and CV tasks, more and more researchers are starting to explore the image-language, and video-language tasks \cite{vinyals2015show, lee2018stacked, antol2015vqa, tapaswi2016movieqa}. The latter is more related to our task since it requires exploring the information in the temporal dimension. Many attempts \cite{sun2019videobert, li2020hero, lei2021less, tang2021decembert} have been done on video-language following a pretraining then finetuning manner. They first adopt some proxy tasks to train the model in an 
self-supervised manner, categorized into completion, matching and ordering. Then, the well-learned representations should be transferred to downstream tasks, e.g. text-based video retrieval \cite{yu2018joint}, action step localization \cite{zhukov2019cross}, video question answering \cite{tapaswi2016movieqa}. More details about vision-language learning tasks can be found in the survey \cite{ruan2021survey}. 

Tasks mentioned above usually make video-level or frames-level prediction and do not require fine-grained features. In contrast, text-based video segmentation requires to predict on pixel-level. So pre-trained video-language models can not be directly applied to our task.

\noindent\textbf{Vision Transformer} % ️️✔️
Vaswani \etal \cite{vaswani2017attention} first propose the transformer, which shows its predominance in many Natual Language Processing (NLP) tasks. The main component of transformers is the self-attention mechanism, which can model long-range dependencies in the data. Computer vision community views this advantage and attempt to design transformer-based models for image classification \cite{dosovitskiy2020image, Yang2020Distill, liu2021swin, yuan2021tokens, metaformer}, object detection \cite{carion2020end, zhu2020deformable} and video understanding \cite{arnab2021vivit, bertasius2021space}. Transformers have also been introduced into some multi-modal tasks. Hu \etal \cite{hu2021unit} propose a unified transformer model jointly trained on multiple tasks, including not only vision-only and language-only tasks but also vision-and-language reasoning. Chen \etal \cite{chen2021mm} adopt a multi-modal video transformer to collaboratively fuse appearance, motion and audio features for video action recognition. Liu \etal \cite{liu2021visual} adopt two transformers to extract appearance and depth information for saliency detection.

In this paper, we propose a transformer-based module that contains cross-modal attention and temporal attention. The former incorporates motion modalities with appearance and linguistic features, and the latter focuses on aggregating temporal information.

\vspace{-3mm}
\section{Method} 
\begin{figure*}[!t]
  \centering
  \includegraphics[width=0.95\linewidth]{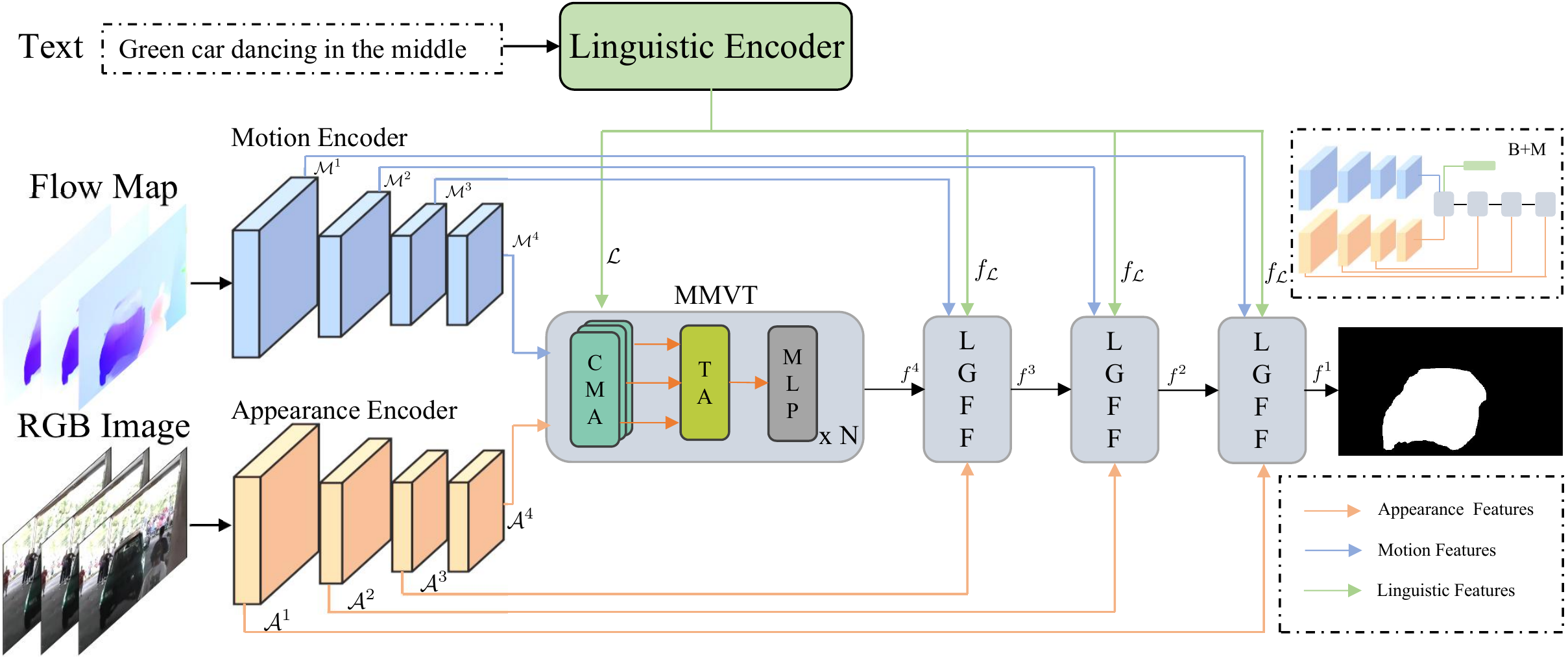}% ️️✔️
  \caption{Overview of the proposed model. MMVT: Multi-modal video transformer. CMA: Cross-modal attention. TA: Temporal attention. LGFF: Language-guided feature fusion. "B+M" is the baseline model with motion information, details about which can be found in \ref{albation study}. Here, we do not show the multi-modal alignment loss for simplification.}
  \label{network overview}
  \vspace{-6mm}
\end{figure*}

\vspace{-2mm}
The overall architecture of the proposed method is shown in Figure~\ref{network overview}. For a video sequence, we have $T$ frames, their corresponding flow maps, and the text, which describe the target object and its action. First, we adopt three encoders to extract appearance, motion, and language features, respectively. Then, the extracted three kinds of high-level features will be concatenated together and input into our multi-modal video transformer (MMVT) to fuse cross-modal features and build temporal relationships between frames. In the decoder, appearance and motion features from different levels will be progressively fused with language features in our language-guided feature fusion module (LGFF) and predict the final segmentation mask. During training, a multi-modal alignment loss is added to align features from different modalities. In the following paper, we will first simply introduce features extraction encoders in Section~\ref{Encoders}, then illustrate detailedly the proposed MMT, LGFF, MMAL in Section~\ref{MMVT}, \ref{LGFF}, and \ref{MMAL}, respectively.

\vspace{-2mm}
\subsection{Encoders} \label{Encoders}
\vspace{-2mm}
We adopt two visual backbones for a video clip with its flow maps to extract the multi-level appearance features $\mathcal{A}^{i} \in \mathbb{R}^{T \times C^{i}_{\mathcal{A}} \times H^{i} \times W^{i}}$ and motion features $\mathcal{M}^{i} \in \mathbb{R}^{T \times C^{i}_{\mathcal{M}} \times H^{i} \times W^{i}}$, where $i\in [1,4]$ denotes the $i$th stage from the backbone. Following \cite{bellver2020refvos}, we leverage the bidirectional transformer model BERT \cite{devlin2018bert} as the linguistic encoder to extract linguistic features. Specifically, we first tokenize the text and add the [CLS] and [SEP] tokens at the beginning and end of the tokenized sequence. Then we feed the tokens sequence into BERT and obtain the token representations as linguistic feature $\mathcal{L} \in  \mathbb{R}^{L \times C_{\mathcal{L}}}$. 

% ️️✔️
We adopt a 1D convolution layer for the linguistic feature to reduce its channel dimension to $C$, obtaining $z_{\mathcal{L}} \in  \mathbb{R}^{L \times C}$. For the high-level appearance feature $\mathcal{A}^{4}$ and motion feature $\mathcal{M}^{4}$, we first respectively concatenate an 8-dimensional coordinate feature $PC^{4} \in \mathbb{R}^{8 \times H^{4} \times W^{4}}$ with them like \cite{wang2019asymmetric} to encode the spatial location information. Then, two ASPP modules \cite{chen2017rethinking} are adopted to unify their channel dimensions to $C$, respectively. Finally, two features are flattened and reshaped, resulting  $z_{\mathcal{A}} \in \mathbb{R}^{T  \times H^{4}W^{4} \times C}$ and $z_{\mathcal{M}} \in \mathbb{R}^{T  \times H^{4}W^{4} \times C}$, respectively.

\vspace{-2mm}
\subsection{Multi-Modal Video Transformer} \label{MMVT}
\vspace{-2mm}
As discussed in Section~\ref{Introduction}, to explore the rich multi-modal interaction and leverage temporal information in different frames, we propose our Multi-Modal Video Transformer (MMVT). From Figure~\ref{network overview}, each layer of our MMVT contains three components: cross-modal attention (CMA), temporal attention (TA), and MLP. The multi-layer perceptron block (MLP) is a common component in transformers \eg \cite{vaswani2017attention, dosovitskiy2020image} and we do not talk about it here.

% ️️✔️
In our cross-modal attention module, we aim to promote the interaction between different modalities in a single frame. Based on this, we first concatenate high-level features from three modalities and obtain feature $z \in \mathbb{R}^{T \times (2HW + L) \times C}$. Here,  we omit the superscript of $H$ and $W$ for simplification. These can be formulated as:
\begin{equation}
\vspace{-1mm}
z = Cat(z_{\mathcal{A}}, z_{\mathcal{M}}, z_{\mathcal{L}}).
\vspace{-1mm}
\end{equation}

We omit the broadcast operation along the temporal dimension for $z_{\mathcal{L}}$ here. Then, we pass it through a layer normalization (LN) \cite{ba2016layer} and we input $z$ into the multi-head self attention (MSA) \cite{vaswani2017attention}. Note that, a residual connection is added here to improve robustness. Formally, this can be defined as:
\vspace{-1mm}
\begin{equation}
\vspace{-1mm}
z' = MSA(LN(z)) + z.
\end{equation}
This process acts along the temporal dimension so that multi-model features in every frame can be well fused. % ️️✔️

In our temporal attention module, the fused multi-modal features from different frames can interact with each other. First, we chunk $z'$ into $z'_{\mathcal{A}}  \in \mathbb{R}^{T  \times HW \times C} $, $z'_{\mathcal{M}}  \in \mathbb{R}^{T  \times HW \times C}$, $z'_{\mathcal{L}} \in \mathbb{R}^{T  \times L \times C}$. Here, $z'_{\mathcal{A}}$ can be considered as the appearance feature that has been enhanced by other modal features. To reduce the computational complexity, we only build the temporal relationships for $z'_{\mathcal{A}}$. Before feeding into MSA, we first flatten $z'_{\mathcal{A}}$ into $\mathbb{R}^{THW \times C}$, so that the information in temporal dimension can participate in the interaction. Then it can be formulated as follow:
% This can be formulated as:

% \begin{equation}
% [z'_{\mathcal{A}}, z'_{\mathcal{M}}, z'_{\mathcal{L}}] = split(z').
% \end{equation}

\vspace{-2mm}
\begin{equation}
z''_{\mathcal{A}} =MSA(LN(z'_{\mathcal{A}})) + z'_{\mathcal{A}}
\end{equation}
After that, $z''_{\mathcal{A}}$ is reshaped back to $\mathbb{R}^{T \times HW \times C}$. 

By doing the process above, the information contained in a frame can flow to other frames. After that, we concatenate the feature $z''_{\mathcal{A}}$, which has been enhanced by other frames, with $z'_{\mathcal{M}}$ and $z'_{\mathcal{L}}$, resulting in $z''$. Finally, we adopt the MLP to increase nonlinearity. These can be formulated as:
\vspace{-2mm}
\begin{equation}
z'' = Cat(z''_{\mathcal{A}}, z'_{\mathcal{M}}, z'_{\mathcal{L}}),
\vspace{-2mm}
\end{equation}
\vspace{-4mm}
\begin{equation}
\vspace{-1mm}
z''' = MLP(z'') + z''.
\end{equation}
% where 
% \begin{equation}
% z''' = [z'''_{\mathcal{A}}, z'''_{\mathcal{M}}, z'''_{\mathcal{L}}]
% \end{equation}

Since $z'_{\mathcal{A}}$ already contains information from other modalities, the multi-modal information can exchange and fuse between frames via the interaction of $z'_{\mathcal{A}}$ in the temporal attention module. Note that, these are all processes in one layer of MMVT. By stacking them for several layers, multi-modal features from different frames can be well fused and aggregated. Here, we set the number of layers to four by default. % ️️✔️

\begin{figure}[!t]
    \centering
    \includegraphics[width=1\linewidth]{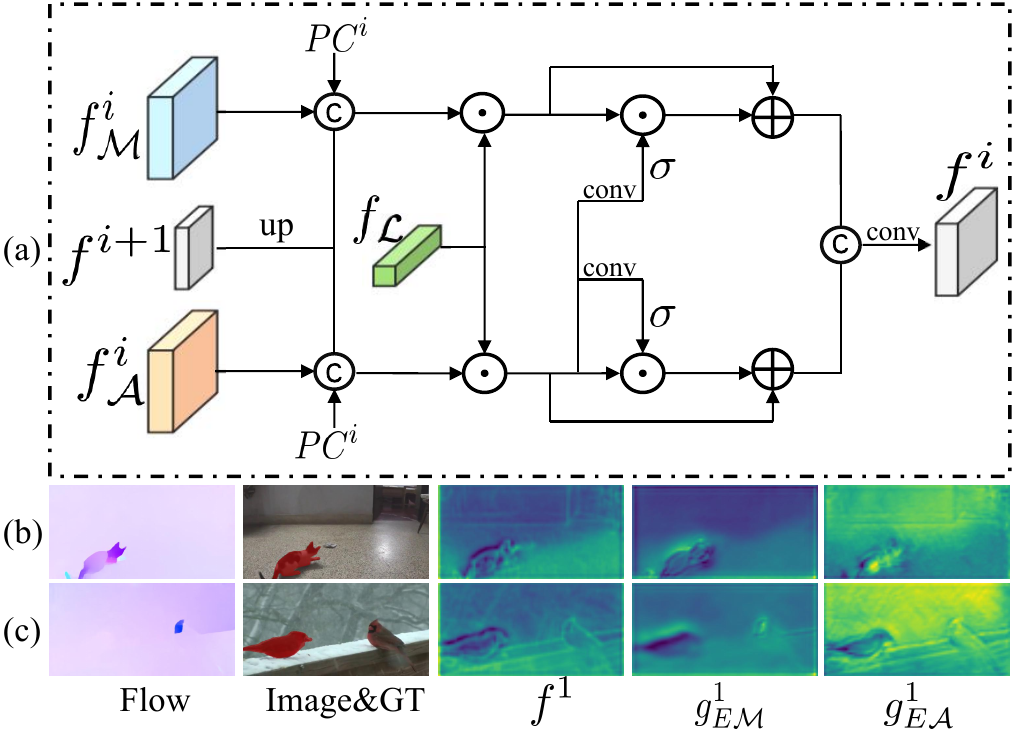}
    \caption{(a) Language-Guided Feature Fusion Module. "up": Upsample operation. $PC^{i}$: Coordinate feature for $i$th level. "C": Concatenation operation. $\odot$: Element-wise Multiplication.  $\oplus$: Element-wise Addition. (b)(c) We visualize the feature map $g_{E\mathcal{M}}^1$, $g_{E\mathcal{A}}^1$ and $f^{1}$.}  % ️️✔️
    \label{LGFF_figure}
    \vspace{-0.4cm}
\end{figure}

\subsection{Language-Guided Feature Fusion Module} \label{LGFF}
\vspace{-2mm}
Our language-guided feature fusion module (LGFF) aims at progressively fusing multi-modal features from different feature levels. As illustrated in Figure~\ref{LGFF_figure} (a), we first adopt two $1 \times 1$ convolution layers to reduce the channel number of appearance feature $f_{\mathcal{A}}^{i}$ and motion feature $f_{\mathcal{M}}^{i}$ to $C$. Then, each feature will be concatenated with the feature from the previous LGFF module $f^{i+1}$ and the 8-dimensional coordinate feature, followed by a $3 \times 3$ convolution layer to fuse them. The feature $f^{i+1}$ contains higher-level and semantically stronger information, while the coordinate feature can provide spatial location information.

Then, we need to emphasize the important region in the feature map with the guidance from linguistic features. Since [CLS] token in $\mathcal{L}$ has aggregated the representation of the whole sentence \cite{devlin2018bert}, we multiply it with two fused features, and obtain the enhanced appearance and motion feature, respectively. We can formulate this process as:
\begin{equation}
\vspace{-1mm}
f_{E\mathcal{A}}^i = f_{\mathcal{L}} \odot Conv_{3}([PC^{i}, Up(f^{i+1}), f_{\mathcal{A}}^{i}]),
\end{equation}
\begin{equation}
f_{E\mathcal{M}}^i = f_{\mathcal{L}} \odot Conv_{3}([PC^{i}, Up(f^{i+1}), f_{\mathcal{M}}^{i}]).
\end{equation}
Here, $\odot$ denotes element-wise multiplication. $f_{\mathcal{L}}$ represents the [CLS] token in linguistic features $\mathcal{L}$. Through this process, the region related to the text in the feature will be selected and emphasized.

Since appearance usually contains more information than motion features, we adopt $f_{EA}^i$ to generate two spatial-attention maps $att_{\mathcal{A}}$ and  $att_{\mathcal{M}}$ through a $1 \times 1$ convolution layer followed by a sigmoid function to further emphasize the target region. Note that two convolution layers here do not share parameters. The residual connect is adopted here to avoid losing some meaningful information.
\vspace{-2mm}
\begin{equation}
g_{E\mathcal{A}}^i = att_{\mathcal{A}} \odot f_{E\mathcal{A}}^i  + f_{E\mathcal{A}}^i,
\end{equation}
\vspace{-4mm}
\begin{equation}
g_{E\mathcal{M}}^i  = att_{\mathcal{M}} \odot f_{E\mathcal{M}}^i  + f_{E\mathcal{M}}^i,
\end{equation}
where $g_{E\mathcal{A}}^i$ and $g_{E\mathcal{M}}^i$ are two obtained features.

Finally, they are concatenated together and further fused with two $3 \times 3$ convolution layers with the ReLU function, resulting in $f^{i}$.
We insert three LGFF modules into our network as the decoder, hence $i \in [1, 3]$. Note that we adopt $z'''_{\mathcal{A}}$ as the feature from the highest level $f^{4}$ in the first LGFF module since it has been enhanced by other modalities and temporal information from different frames in MMVT. From Figure~\ref{LGFF_figure} (b)(c), we can find that, no matter whether the flow map can highlight the target object or not, $g_{E\mathcal{M}}^i$ can distinguish the target object from other regions and incorporate well with $g_{E\mathcal{A}}^i$ to generate the final output feature $f^{1}$.

\begin{figure}[!t]
    \centering
    \includegraphics[width=1\linewidth]{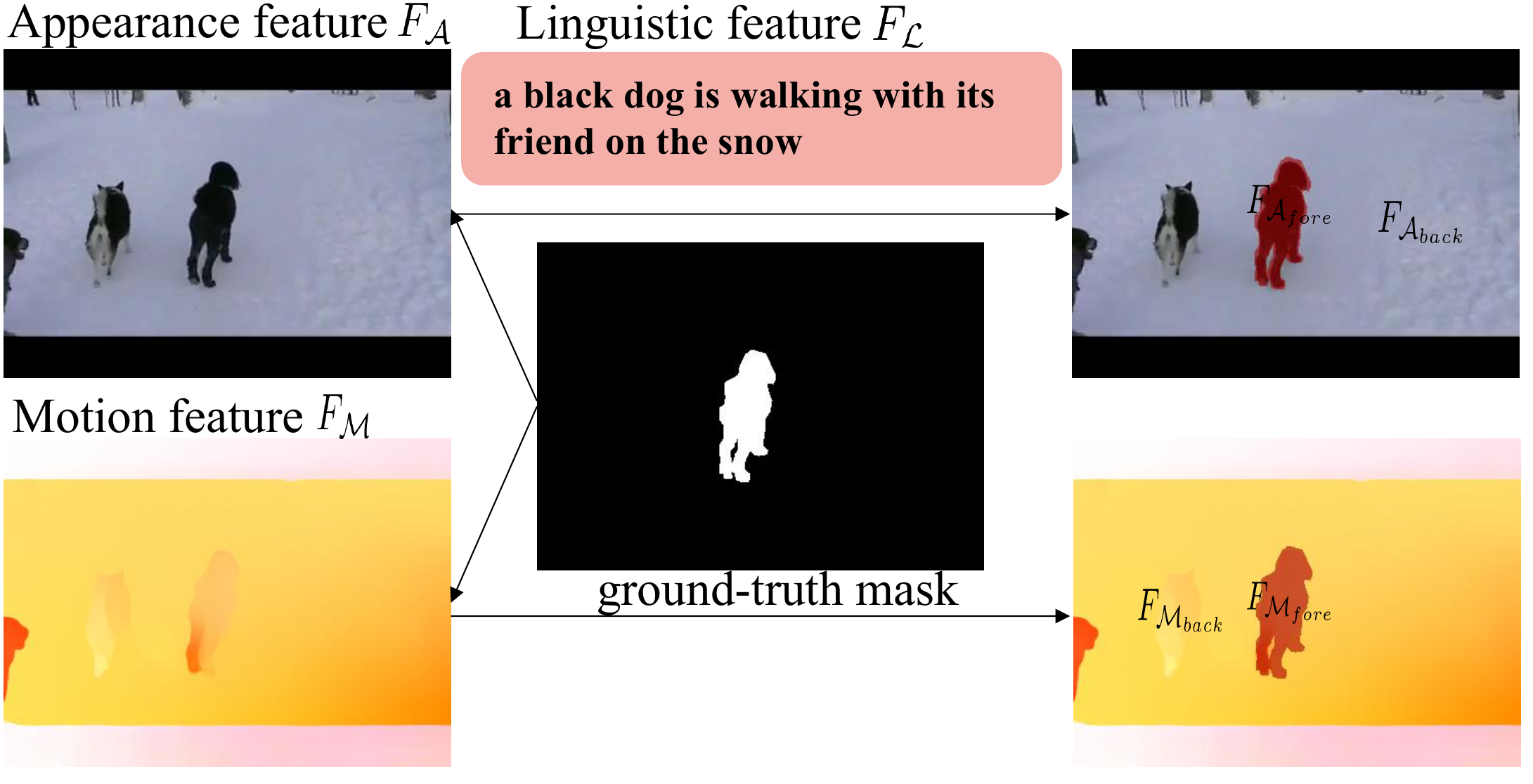}
    \caption{We adopt the ground-truth mask to distinguish the features that belong to the foreground or the background in appearance features$f_{\mathcal{A}}$ and motion feature $f_{\mathcal{M}}$.}  % ️️✔️
    \label{alignmentlss_figure}
    \vspace{-0.4cm}
\end{figure}

\subsection{Multi-modal Alignment Loss} \label{MMAL}
\vspace{-2mm}
Although our model has achieved good performance with the two modules above, we notice that there may exist some semantic gap between the three modalities features since they are extracted from encoders that are pre-trained on different source data \cite{huang2020pixel}. Based on this, we propose our multi-modal alignment loss so that three modalities features can be explicitly aligned.  % ️️✔️

Specifically, we consider that the features belonging to the target object from appearance features are foreground features, and other features are background features. For motion features, it can also be categorized into the foreground and background features. Then, the feature alignment rules are defined as (1) the linguistic features should be close to foreground features from both appearance and motion features in embedding space, while far away from background features. (2) Appearance and motion features from the same category should be close to each other, meanwhile far away from other category features. Since the multi-modal alignment loss is defined per frame, we do not need to consider the temporal dimension here.  % ️️✔️

First, for each frame, we obtain a whole representation of appearance feature $F_{\mathcal{A}}$ by upsampling $f_{E\mathcal{A}}^2$, $f_{E\mathcal{A}}^3$ and concatenating them with $f_{E\mathcal{A}}^1$ together. $F_{\mathcal{M}}$ is obtained in the same way. Here we also adopt the [CLS] token $F_\mathcal{L}$ in $\mathcal{L}$ as the whole representation of the text. We employ three MLP functions to transform $F_{\mathcal{A}}$,  $F_{\mathcal{M}}$, and $F_{\mathcal{L}}$ into the same embedding space with the same channel numbers $c$.  % ️️✔️

Now, we need to distinguish the feature belongs to the target object from other features in $F_{\mathcal{A}}$ and $F_{\mathcal{M}}$. This can be easily realized by leveraging the ground-truth mask. For example, in Figure~\ref{alignmentlss_figure}, we can obtain foreground features $F_{\mathcal{A}_{fore}}$ and background features $F_{\mathcal{A}_{back}}$, where we know that $F_{\mathcal{A}_{fore}} \cup F_{\mathcal{A}_{back}} = F_{\mathcal{A}}$. We can obtain the alignment score $p_{\mathcal{AL}}$ between each element $f_{\mathcal{A}}^{i} \in F_{\mathcal{A}}$ and $F_{\mathcal{L}}$ by:
\vspace{-2mm}
\begin{equation}
\hat{p}_{\mathcal{AL}}^{i} = \sigma(\tan (\frac{\pi}{2} sim(f_{\mathcal{A}}^{i}, F_{\mathcal{L}}))), 
\vspace{-2mm}
\end{equation}

where $sim$ represents the function to calculate the cosine similarity. If $f_{\mathcal{A}}^{i}$ is close to $ F_{\mathcal{L}}$ in the embedding space, their cosine similarity will be close to $1$ then the alignment score $\hat{p}_{\mathcal{AL}}^{i}$ will be close to $1$, otherwise $\hat{p}_{\mathcal{AL}}^{i}$ will be close to $0$. Based on this, we can define its label $p_{\mathcal{AL}}^{i}$ as: if $f_{\mathcal{A}}^{i} \in F_{\mathcal{A}_{fore}}$, $p_{\mathcal{AL}}^{i}=1$ otherwise $p_{\mathcal{AL}}^{i}=0$. Now the alignment loss $L_{\mathcal{AL}}$ between $F_{\mathcal{A}}$ and $F_{\mathcal{L}}$ can be defined as:
\vspace{-1mm}
\begin{equation}
L_{\mathcal{AL}} = -\sum p_{\mathcal{AL}}^{i} \log \hat{p}_{\mathcal{AL}}^{i} + (1-p_{\mathcal{AL}}^{i}) \log(1- \hat{p}_{\mathcal{AL}}^{i}).
\end{equation}
The alignment loss $L_{\mathcal{ML}}$ between $F_{\mathcal{M}}$ and $F_{\mathcal{L}}$ can also be defined in the same way.

For appearance features $f_{\mathcal{A}}^{i} \in F_{\mathcal{A}}$ and motion features $f_{\mathcal{M}}^{i} \in F_{\mathcal{M}}$, we can also align them together. The alignment score can be defined as:
\vspace{-2mm}
\begin{equation} 
\hat{p}_{\mathcal{AM}}^{i,j} = \sigma(\tan (\frac{\pi}{2} sim(f_{\mathcal{A}}^{i}, f_{\mathcal{M}}^{j}))).
\vspace{-2mm}
\end{equation}

When $f_{\mathcal{A}}^{i}$ and $F_{\mathcal{M}}^{j}$ belongs to the foreground or background at the same time, its label $p_{\mathcal{AM}}^{i,j} =1 $, otherwise  $p_{\mathcal{AM}}^{i,j}=0$. The alignment loss $L_{\mathcal{AM}}$ can be defined as:
\begin{equation}
L_{\mathcal{AM}} = -\sum p_{\mathcal{AM}}^{i,j} \log \hat{p}_{\mathcal{AM}}^{i,j} + (1-p_{\mathcal{AM}}^{i,j}) \log(1- \hat{p}_{\mathcal{AM}}^{i,j}).
\end{equation}
Finally, we define the multi-modal alignment loss as:
\begin{equation}
L^{align} =L_{\mathcal{AL}} + L_{\mathcal{ML}} + L_{\mathcal{AM}}.
\end{equation} % ️️✔️

\vspace{-4mm}
\section{Experiments}
\vspace{-1mm}  
\subsection{Datasets and Evaluation Metrics}
\vspace{-2mm}
Following prior works, we conduct experiments on two popular text-based video segmentation datasets including  \textbf{A2D Sentences} \cite{gavrilyuk2018actor} and \textbf{J-HMDB} Sentences \cite{gavrilyuk2018actor}. These two datasets are extended by Gavrilyuk \etal \cite{gavrilyuk2018actor} via providing a referring language for each target object in Actor-Action Dataset (A2D) \cite{xu2015can} and J-HMDB \cite{jhuang2013towards}.

\textbf{A2D Sentences} contains 3,782 videos, which are split into 3,036 and 746 videos for training and testing, respectively. There are 3 to 5 frames with pixel-level annotations in each video for training and evaluating segmentation performance. Besides, there are 6,655 sentences to describe the actors and their actions in each video. \textbf{J-HMDB Sentences} contains 928 videos from 21 action classes with corresponding 928 sentences. All frames in it are annotated at the pixel level. Previous methods usually evaluate their generalization ability on this dataset.

Intersection-over-union (IoU) is the ratio of intersection area over union area between the ground-truth mask and prediction. Following prior works, we adopt \textbf{Overall IoU} and \textbf{Mean IoU} to evaluate the performance. The former treats ground-truth masks and predictions on the testing dataset as a whole, resulting in favor of larger objects, while the latter is the averaged IoU overall test samples. We also adopt \textbf{P@X} to measure the percentage of samples whose IoU are higher than the threshold X, where X $\in [0.5, 0.6, 0.7, 0.8, 0.9]$. The mean average precision (\textbf{mAP}) over 0.5:0.95 is also adopted.

\vspace{-2mm}  
\subsection{Implementation Details}
\vspace{-2mm}
Following \cite{li2019motion}, we adopt the ResNet-101 and ResNet-34 \cite{he2016deep} as the appearance and motion encoders to extract appearance and motion features. The stride of four stages in two encoders is set as 2, 2, 2, and 1, respectively. RAFT \cite{teed2020raft} is employed to generate optical flow maps. We adopt an Adam \cite{kingma2014adam} optimizer with the learning rate $2 \times 10^{-5}$ to train the whole network. The batch size is set to 8, and each batch contains a video clip with three frames. We set the maximum training step to 30,000, and the learning rate is divided by ten at 25,000 and 28,000, respectively. Following the settings in prior works, all frames are resized and padded to $320 \times 320$. The maximum length of each input sentence is 20. All experiments are conducted on 2 NVIDIA Tesla V100 GPUs. 

\begin{table*}[t!]
	\centering
	\scriptsize
	\renewcommand{\arraystretch}{1.0}
	\renewcommand{\tabcolsep}{4.3mm}
	\caption{Comparison with state-of-the-art methods on A2D Sentences testing set. $\dagger$ denotes adopting additional optical flow input.}
	\vspace{-3mm}
	\begin{tabular}{cccc|cc|ccccc|c|cc|} 
	   \hline
        \multicolumn{4}{|c|}{\multirow{2}{*}{\textbf{Methods}}}  & \multicolumn{2}{|c|}{\multirow{2}{*}{\textbf{Venue}}} & \multicolumn{5}{c|}{\textbf{Precision}}  & \multicolumn{1}{c|}{\textbf{mAP}}  & \multicolumn{2}{c|}{\textbf{IoU}} \\ \cline{7-14}
        \multicolumn{4}{|c|}{}                                   &  \multicolumn{2}{|c|}{}                                 &\multicolumn{1}{c|}{P@0.5}               &\multicolumn{1}{c|}{P@0.6} &\multicolumn{1}{c|}{P@0.7}      &\multicolumn{1}{c|}{P@0.8}  &\multicolumn{1}{|c|}{P@0.9}  &\multicolumn{1}{c|}{0.5:0.95}      &\multicolumn{1}{c|}{Overall}  &\multicolumn{1}{c|}{Mean}   \\  \hline
        
	    \multicolumn{4}{|c|}{Hu \etal \cite{hu2016segmentation}}     & \multicolumn{2}{c|}{ECCV2016}                  & \multicolumn{1}{c|}{34.8}  & \multicolumn{1}{c|}{23.6}  & \multicolumn{1}{c|}{13.3} & \multicolumn{1}{c|}{3.3 } & \multicolumn{1}{c|}{0.1} & 13.2 & \multicolumn{1}{c|}{47.4} & 35.0 \\
		\multicolumn{4}{|c|}{Li \etal \cite{li2017tracking}}               & \multicolumn{2}{c|}{CVPR2017}           & \multicolumn{1}{c|}{38.7}  & \multicolumn{1}{c|}{29.0}  & \multicolumn{1}{c|}{17.5} & \multicolumn{1}{c|}{6.6 } & \multicolumn{1}{c|}{0.1} & 16.3 & \multicolumn{1}{c|}{51.5} & 35.4 \\
		\multicolumn{4}{|c|}{Gavrilyuk \etal \cite{gavrilyuk2018actor}}           & \multicolumn{2}{c|}{CVPR2018}       & \multicolumn{1}{c|}{47.5}  & \multicolumn{1}{c|}{34.7}  & \multicolumn{1}{c|}{21.1} & \multicolumn{1}{c|}{8.0 } & \multicolumn{1}{c|}{0.2} & 19.8 & \multicolumn{1}{c|}{53.6} & 42.1 \\
		\multicolumn{4}{|c|}{Gavrilyuk \etal $\dagger$ \cite{gavrilyuk2018actor}}  & \multicolumn{2}{c|}{CVPR2018}      & \multicolumn{1}{c|}{50.0}  & \multicolumn{1}{c|}{37.6}  & \multicolumn{1}{c|}{23.1} & \multicolumn{1}{c|}{9.4 } & \multicolumn{1}{c|}{0.4} & 21.5 & \multicolumn{1}{c|}{55.1} & 42.6 \\
		\multicolumn{4}{|c|}{ACGA \cite{wang2019asymmetric}}                        & \multicolumn{2}{c|}{ICCV2019}      & \multicolumn{1}{c|}{55.7}  & \multicolumn{1}{c|}{45.9}  & \multicolumn{1}{c|}{31.9} & \multicolumn{1}{c|}{16.0} & \multicolumn{1}{c|}{2.0} & 27.4 & \multicolumn{1}{c|}{60.1} & 49.0  \\ 
	    \multicolumn{4}{|c|}{VT-Capsule \cite{mcintosh2020visual}}                  & \multicolumn{2}{c|}{CVPR2020}      & \multicolumn{1}{c|}{52.6}  & \multicolumn{1}{c|}{45.0}  & \multicolumn{1}{c|}{34.5} & \multicolumn{1}{c|}{20.7} & \multicolumn{1}{c|}{3.6} & 30.3 & \multicolumn{1}{c|}{56.8} & 46.0 \\ 
		\multicolumn{4}{|c|}{CMDY \cite{wang2020context}}                        & \multicolumn{2}{c|}{AAAI2020}      & \multicolumn{1}{c|}{60.7}  & \multicolumn{1}{c|}{52.5}  & \multicolumn{1}{c|}{40.5} & \multicolumn{1}{c|}{23.5} & \multicolumn{1}{c|}{4.5} & 33.3 & \multicolumn{1}{c|}{62.3} & 53.1 \\

	    \multicolumn{4}{|c|}{PRPE \cite{ning2020polar}}                        & \multicolumn{2}{c|}{IJCAI2020}       & \multicolumn{1}{c|}{63.4}      & \multicolumn{1}{c|}{57.9}      & \multicolumn{1}{c|}{48.3}     & \multicolumn{1}{c|}{32.2}     & \multicolumn{1}{c|}{8.3}    & \multicolumn{1}{c|}{38.8} & \multicolumn{1}{c|}{66.1} & \multicolumn{1}{c|}{52.9} \\

		\multicolumn{4}{|c|}{CSTM \cite{hui2021collaborative}}                        & \multicolumn{2}{c|}{CVPR2021}      & \multicolumn{1}{c|}{\textbf{65.4}}  & \multicolumn{1}{c|}{58.9}  & \multicolumn{1}{c|}{49.7} & \multicolumn{1}{c|}{33.3} & \multicolumn{1}{c|}{9.1} & 39.9 & \multicolumn{1}{c|}{66.2} & \multicolumn{1}{c|}{\textbf{56.1}} \\\hline

	    \multicolumn{4}{|c|}{Our $\dagger$}                        & \multicolumn{2}{c|}{--}       & \multicolumn{1}{c|}{64.5}      & \multicolumn{1}{c|}{\textbf{59.7}}      & \multicolumn{1}{c|}{\textbf{52.3}}     & \multicolumn{1}{c|}{\textbf{37.5}}     & \multicolumn{1}{c|}{\textbf{13.0}} & \multicolumn{1}{c|}{\textbf{41.9}} & \multicolumn{1}{c|}{\textbf{67.3}} & \multicolumn{1}{c|}{55.8}  \\ \hline
\vspace{-6mm}
	\end{tabular}
	\label{A2D}
\end{table*}

\begin{table*}[t!]
	\centering
	\scriptsize
	\renewcommand{\arraystretch}{1.0}
	\renewcommand{\tabcolsep}{4.3mm}
	\caption{Comparison with state-of-the-art methods on J-HMDB Sentences testing set. All methods adopt the best model trained on A2D Sentences to directly eval on J-HMDB Sentences without finetuning. $\dagger$ denotes adopting additional optical flow input.}
\vspace{-3mm}
	\begin{tabular}{cccc|cc|ccccc|c|cc|} 
	   \hline
        \multicolumn{4}{|c|}{\multirow{2}{*}{\textbf{Methods}}}  & \multicolumn{2}{|c|}{\multirow{2}{*}{\textbf{Venue}}} & \multicolumn{5}{c|}{\textbf{Precision}}  & \multicolumn{1}{c|}{\textbf{mAP}}  & \multicolumn{2}{c|}{\textbf{IoU}} \\ \cline{7-14}
        \multicolumn{4}{|c|}{}                                   &  \multicolumn{2}{|c|}{}                                 &\multicolumn{1}{c|}{P@0.5}               &\multicolumn{1}{c|}{P@0.6} &\multicolumn{1}{c|}{P@0.7}      &\multicolumn{1}{c|}{P@0.8}  &\multicolumn{1}{|c|}{P@0.9}  &\multicolumn{1}{c|}{0.5:0.95}      &\multicolumn{1}{c|}{Overall}  &\multicolumn{1}{c|}{Mean}   \\  \hline
        
	    \multicolumn{4}{|c|}{Hu \etal \cite{hu2016segmentation}}     & \multicolumn{2}{c|}{ECCV2016}                & \multicolumn{1}{|c|}{63.3} & \multicolumn{1}{c|}{35.0}  & \multicolumn{1}{c|}{8.5} & \multicolumn{1}{c|}{0.2} & \multicolumn{1}{c|}{0.0} &  \multicolumn{1}{c|}{17.8} & \multicolumn{1}{c|}{54.6} & \multicolumn{1}{c|}{52.8} \\

		\multicolumn{4}{|c|}{Li \etal \cite{li2017tracking}}             & \multicolumn{2}{c|}{CVPR2017}    & \multicolumn{1}{c|}{57.8}                & \multicolumn{1}{|c|}{33.5} & \multicolumn{1}{c|}{10.3}  & \multicolumn{1}{c|}{0.6} & \multicolumn{1}{c|}{0.0} & \multicolumn{1}{c|}{17.3} &  \multicolumn{1}{c|}{52.9} & \multicolumn{1}{c|}{49.1}  \\

		\multicolumn{4}{|c|}{Gavrilyuk \etal \cite{gavrilyuk2018actor}}             & \multicolumn{2}{c|}{CVPR2018}     & \multicolumn{1}{c|}{69.9}                & \multicolumn{1}{|c|}{46.0} & \multicolumn{1}{c|}{17.3}  & \multicolumn{1}{c|}{1.4} & \multicolumn{1}{c|}{0.0} & \multicolumn{1}{c|}{23.3} & \multicolumn{1}{c|}{54.1}  & \multicolumn{1}{c|}{54.2}  \\

% 		\multicolumn{4}{|c|}{Gavrilyuk \etal $\dagger$ \cite{gavrilyuk2018actor}}   & \multicolumn{2}{c|}{CVPR2018}                & \multicolumn{1}{|c|}{71.2} & \multicolumn{1}{c|}{51.8}  & \multicolumn{1}{c|}{26.4} & \multicolumn{1}{c|}{3.0} & \multicolumn{1}{c|}{0.0} & \multicolumn{1}{c|}{26.7} & \multicolumn{1}{c|}{55.5} & \multicolumn{1}{c|}{57.0} \\

		\multicolumn{4}{|c|}{ACGA \cite{wang2019asymmetric}}                       & \multicolumn{2}{c|}{ICCV2019}                & \multicolumn{1}{|c|}{75.6} & \multicolumn{1}{c|}{56.4}  & \multicolumn{1}{c|}{28.7} & \multicolumn{1}{c|}{3.4} & \multicolumn{1}{c|}{0.0} &  \multicolumn{1}{c|}{28.9} & \multicolumn{1}{c|}{57.6} & \multicolumn{1}{c|}{58.4} \\

	    \multicolumn{4}{|c|}{VT-Capsule \cite{mcintosh2020visual}}                  & \multicolumn{2}{c|}{CVPR2020}                & \multicolumn{1}{|c|}{67.7} & \multicolumn{1}{c|}{51.3}  & \multicolumn{1}{c|}{28.3} & \multicolumn{1}{c|}{5.1} & \multicolumn{1}{c|}{0.0} & \multicolumn{1}{c|}{26.1}  & \multicolumn{1}{c|}{53.5} & \multicolumn{1}{c|}{55.0}  \\
		
		\multicolumn{4}{|c|}{CMDY \cite{wang2020context}}                        & \multicolumn{2}{c|}{AAAI2020}                & \multicolumn{1}{|c|}{74.2} & \multicolumn{1}{c|}{58.7}  & \multicolumn{1}{c|}{31.6} & \multicolumn{1}{c|}{4.7} & \multicolumn{1}{c|}{0.0}  & \multicolumn{1}{c|}{30.1} & \multicolumn{1}{c|}{55.4} & \multicolumn{1}{c|}{57.6} \\
		
	    \multicolumn{4}{|c|}{PRPE \cite{ning2020polar}}                        & \multicolumn{2}{c|}{IJCAI2020}       & \multicolumn{1}{c|}{69.1}      & \multicolumn{1}{c|}{57.2}      & \multicolumn{1}{c|}{31.9}     & \multicolumn{1}{c|}{6.0}     & \multicolumn{1}{c|}{\textbf{0.1}}    & \multicolumn{1}{c|}{29.4} & \multicolumn{1}{c|}{-} & \multicolumn{1}{c|}{-} \\

		\multicolumn{4}{|c|}{CSTM \cite{hui2021collaborative}}                    & \multicolumn{2}{c|}{CVPR2021}               & \multicolumn{1}{c|}{78.3}                & \multicolumn{1}{|c|}{63.9} & \multicolumn{1}{c|}{37.8}  & \multicolumn{1}{c|}{7.6} & \multicolumn{1}{c|}{0.0} & \multicolumn{1}{c|}{33.5}
		& \multicolumn{1}{c|}{59.8}  &  \multicolumn{1}{c|}{60.4} \\\hline
		
	    \multicolumn{4}{|c|}{Our $\dagger$}                        & \multicolumn{2}{c|}{--}       & \multicolumn{1}{c|}{\textbf{79.9}}      & \multicolumn{1}{c|}{\textbf{71.4}}      & \multicolumn{1}{c|}{\textbf{49.0}}     & \multicolumn{1}{c|}{\textbf{12.6}}     & \multicolumn{1}{c|}{\textbf{0.1}} & \multicolumn{1}{c|}{\textbf{38.6}} & \multicolumn{1}{c|}{\textbf{61.9}} & \multicolumn{1}{c|}{\textbf{61.3}}  \\ \hline
\vspace{-6mm}
	\end{tabular}
	\label{J-HMDB}
\end{table*}

\vspace{-1mm}
\subsection{Comparison with State-of-the-art Methods}
\vspace{-1mm}
\noindent\textbf{A2D Sentences}
We employ the training and testing set of A2D Sentences to train and evaluate our model, respectively. As shown in Table~\ref{A2D}, our method surpass over state-of-the-art method CSTM by 0.8\% , 2.6\% , 4.2\% on Precision @0.6, @0.7 and @0.8, respectively. This means that, when the metric is more stricter, our model can surpass previous methods by a larger margin.  It is noteworthy that, our method achieve 13.0 \% on the most challenging metric Precision @0.9, which means that our method can generate particularly accurate segmentation masks. The mAP and Overall IoU can also be further improved by 2.0\% and 1.1\%, respectively. We also notice that our model is lower than CSTM \cite{hui2021collaborative} by 0.9\% on Precision@0.5, which is because our model tend to generate more accurate and confident results, while some not accurate results from CSTM \cite{hui2021collaborative} can still considered to be True, since the threshold in Precision@0.5 is low. Furthermore, since CSTM generates masks on the feature map with original size while our model predicts on the feature map with $1/4$ original size, they may perform better on small objects. Hence the performance of our method is slightly lower than its on IoU Mean, which treats small objects equally. 

% \vspace{-1mm}
\noindent\textbf{J-HMDB Sentences}
Like previous works, we adopt the J-HMDB Sentences to verify the generalization ability of our method. Following \cite{wang2019asymmetric, hui2021collaborative}, we employ the model that achieves the best performance on A2D Sentences to directly evaluate on the test set of J-HMDB Sentences, which is split by \cite{gavrilyuk2018actor}. As illustrated in Table~\ref{J-HMDB}, our method outperform all previous methods on all metrics. It is easy to find that our model can surpass other methods by a large margin, especially when the metric is strict \eg Precision@0.6, @0.7 and @0.8. This phenomenon is similar to that in A2D Sentences, which means that our model shows more robust performance with the help of well-fused multi-modal information. Note that, like other methods, our approach can not achieve good results on Precision@0.9 (lower than 1 \%), since all methods are not trained or finetuned on J-HMDB Sentences. % ️️✔️

\begin{table*}[t!]
	\centering
	\scriptsize
	\renewcommand{\arraystretch}{1.0}
	\renewcommand{\tabcolsep}{2mm}
	\caption{Quantitative results of each component in our model. Appearance: with appearance feature; Motion: with motion feature; MMVT: Multi-Modal Video Transformer; LGFF: Language-Guided Feature Fusion Module; Align: Multi-modal Alignment Loss.}
\vspace{-3mm}
	\begin{tabular}{cc|cccccccccc|ccccc|c|cc|} 
	   \hline
	                  \multicolumn{2}{|c|}{\multirow{2}{*}{Name}}  &   \multicolumn{10}{|c|}{Settings}  & \multicolumn{5}{c|}{\textbf{Precision}}  & \multicolumn{1}{c|}{\textbf{mAP}}  & \multicolumn{2}{c|}{\textbf{IoU}} \\ 
	                  \cline{11-20} \cline{3-10}
	                  
	               \multicolumn{2}{|c|}{}   & \multicolumn{2}{|c|}{Appearance}  & \multicolumn{2}{|c|}{Motion}  & \multicolumn{2}{|c|}{MMVT}  & \multicolumn{2}{|c|}{LGFF} & \multicolumn{2}{|c|}{Align} & \multicolumn{1}{c|}{P@0.5}               & \multicolumn{1}{c|}{P@0.6} &\multicolumn{1}{c|}{P@0.7}      & \multicolumn{1}{c|}{P@0.8}  &\multicolumn{1}{|c|}{P@0.9}  & \multicolumn{1}{c|}{0.5:0.95}      & \multicolumn{1}{c|}{Overall}  & \multicolumn{1}{c|}{Mean}   \\  \hline
	       
	       	\multicolumn{2}{|c|}{B}  & \multicolumn{2}{|c|}{\ding{51}}    &  \multicolumn{2}{|c|}{}         &  \multicolumn{2}{|c|}{}         &  \multicolumn{2}{|c|}{}         &  \multicolumn{2}{|c|}{}   & \multicolumn{1}{|c|}{55.1}   & \multicolumn{1}{|c|}{50.7}  & \multicolumn{1}{|c|}{44.2}  & \multicolumn{1}{|c|}{31.7}  & \multicolumn{1}{|c|}{9.5}   & \multicolumn{1}{|c|}{35.3}   & \multicolumn{1}{|c|}{61.9}   & \multicolumn{1}{|c|}{48.2} \\
	       	
   	       	\multicolumn{2}{|c|}{B+M}  & \multicolumn{2}{|c|}{\ding{51}}    &  \multicolumn{2}{|c|}{\ding{51}}         &  \multicolumn{2}{|c|}{}         &  \multicolumn{2}{|c|}{}        &  \multicolumn{2}{|c|}{}    & \multicolumn{1}{|c|}{56.8}   & \multicolumn{1}{|c|}{51.9}  & \multicolumn{1}{|c|}{45.0}  & \multicolumn{1}{|c|}{32.3}  & \multicolumn{1}{|c|}{10.0}   & \multicolumn{1}{|c|}{36.3}   & \multicolumn{1}{|c|}{63.5}   & \multicolumn{1}{|c|}{49.5} \\ \hline

 	       	\multicolumn{2}{|c|}{B+T}  & \multicolumn{2}{|c|}{\ding{51}}    &  \multicolumn{2}{|c|}{}         &  \multicolumn{2}{|c|}{\ding{51}}         &  \multicolumn{2}{|c|}{}         &  \multicolumn{2}{|c|}{}   & \multicolumn{1}{|c|}{59.2}   & \multicolumn{1}{|c|}{54.1}  & \multicolumn{1}{|c|}{46.1}  & \multicolumn{1}{|c|}{32.2}  & \multicolumn{1}{|c|}{9.8}   & \multicolumn{1}{|c|}{37.2}   & \multicolumn{1}{|c|}{64.4}   & \multicolumn{1}{|c|}{51.3} \\
 	       	
 	       	\multicolumn{2}{|c|}{B+M+T}  & \multicolumn{2}{|c|}{\ding{51}}    &  \multicolumn{2}{|c|}{\ding{51}}         &  \multicolumn{2}{|c|}{\ding{51}}         &  \multicolumn{2}{|c|}{}         &  \multicolumn{2}{|c|}{}   & \multicolumn{1}{|c|}{62.0}   & \multicolumn{1}{|c|}{56.8}  & \multicolumn{1}{|c|}{48.7}  & \multicolumn{1}{|c|}{34.3}  & \multicolumn{1}{|c|}{10.5}   & \multicolumn{1}{|c|}{39.2}   & \multicolumn{1}{|c|}{64.8}   & \multicolumn{1}{|c|}{53.6} \\ \hline

                	\multicolumn{2}{|c|}{B+T+L}  & \multicolumn{2}{|c|}{\ding{51}}    &  \multicolumn{2}{|c|}{}         &  \multicolumn{2}{|c|}{\ding{51}}         &  \multicolumn{2}{|c|}{\ding{51}}        &  \multicolumn{2}{|c|}{}    & \multicolumn{1}{|c|}{62.0}   & \multicolumn{1}{|c|}{57.4}  & \multicolumn{1}{|c|}{49.6}  & \multicolumn{1}{|c|}{36.2}  & \multicolumn{1}{|c|}{11.6}   & \multicolumn{1}{|c|}{40.1}   & \multicolumn{1}{|c|}{65.5}   & \multicolumn{1}{|c|}{54.0} \\

                	\multicolumn{2}{|c|}{B+M+T+L}  & \multicolumn{2}{|c|}{\ding{51}}    &  \multicolumn{2}{|c|}{\ding{51}}         &  \multicolumn{2}{|c|}{\ding{51}}         &  \multicolumn{2}{|c|}{\ding{51}}       &  \multicolumn{2}{|c|}{}     & \multicolumn{1}{|c|}{63.1}   & \multicolumn{1}{|c|}{58.5}  & \multicolumn{1}{|c|}{51.2}  & \multicolumn{1}{|c|}{37.1}  & \multicolumn{1}{|c|}{12.6}   & \multicolumn{1}{|c|}{41.1}   & \multicolumn{1}{|c|}{66.8}   & \multicolumn{1}{|c|}{54.8} \\\hline

        	\multicolumn{2}{|c|}{B+M+T+L+A}  & \multicolumn{2}{|c|}{\ding{51}}    &  \multicolumn{2}{|c|}{\ding{51}}         &  \multicolumn{2}{|c|}{\ding{51}}         &  \multicolumn{2}{|c|}{\ding{51}}       &  \multicolumn{2}{|c|}{\ding{51}}     & \multicolumn{1}{|c|}{\textbf{64.5}}   & \multicolumn{1}{|c|}{\textbf{59.7}}  & \multicolumn{1}{|c|}{\textbf{52.3}}  & \multicolumn{1}{|c|}{\textbf{37.5}}  & \multicolumn{1}{|c|}{\textbf{13.0}}   & \multicolumn{1}{|c|}{\textbf{41.9}}   & \multicolumn{1}{|c|}{\textbf{67.3}}   & \multicolumn{1}{|c|}{\textbf{55.8}} \\ \hline

	\end{tabular}
	\label{ablation}
	 \vspace{-3mm}
\end{table*}

\begin{figure*}[!t]
  \centering
  \includegraphics[width=1.0\linewidth]{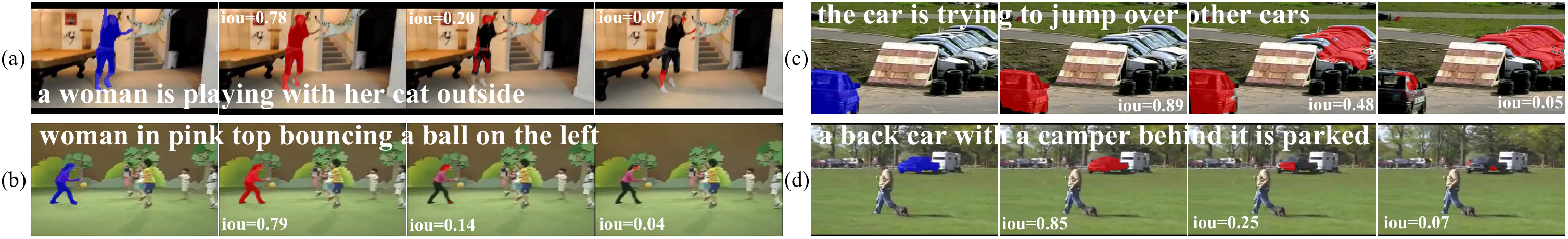}
  \caption{Qualitative results comparison.  From left to right in (a), (b), (c) and (d): ground-truth, "B+M+T+L+A", "B+M", and "B".}
  \label{abl_figure}
  \vspace{-6mm}
\end{figure*}

\vspace{-1mm}
\subsection{Ablation Study} \label{albation study}
\vspace{-1mm}
Following previous works, we conduct ablation experiments on A2D Sentences to thoroughly analyze and verify the effectiveness of the proposed method. 

\vspace{-1mm}
\noindent\textbf{Effectiveness of Each Component.} We first verify each component in our model in Table~\ref{ablation}.  “B+M” is the baseline model shown in Figure~\ref{network overview}, which only adopt concatenation and convolutional layers to fuse multi-modal. In addition, only appearance features are fused in the decoder in "B+M". "B" is the same as "B+M" except without motion branch. By comparing them, we can find that introducing the explicit motion information from optical flow maps can effectively improve the performance. To verify the effectiveness of multi-modal interaction between frames, we replace the concatenation operation in "B+M" with the proposed MMVT and obtain "B+M+T". We find that the performance is significantly improved, especially in mAP and Mean IoU, improved by 2.9\% and 4.1\%, respectively. This benefits from the powerful capacity of fusing multi-modal features between frames in MMVT. Then, we replace all simple concatenation operations in every level of the decoder in "B+M+T" with the proposed LGFF and obtain "B+M+T+L". This demonstrates a remarkable improvement on all metrics, especially in rigorous metrics Precision@0.7, @0.8, and @0.9, which are improved by 2.5\%, 2.7\%, and 2.1\%, respectively. This means the decoder with our LGFF can progressively fuse multi-modal features from different levels and gradually recover the resolution of the feature map, leading to more accurate segmentation masks. Finally, we add the proposed multi-modal alignment loss into "B+M+T+L+A" and the results demonstrate that explicitly aligning multi-modal features can obtain better performance. To further verify the generalization of proposed components, we gradually add our MMVT and LGFF to "B",  resulting in"B+T" and "B+T+L". The results show that only fusing appearance and linguistic features with our MMVT and LGFF can also improve the performance. Hence, our MMVT and LGFF can work well in different settings instead of only handling the setting with motion. % ️️✔️

% \vspace{-1mm}
\noindent\textbf{MMVT Settings.} 
There are two attention modules in our MMVT, named cross model attention (CMA) and temporal attention (TA), respectively. Here, we conduct experiments to verify their effectiveness in Table~\ref{transformer ablation}. We remove all TA modules in MMVT from "B+M+T", which is denoted as "+CMA". We can see that the performance drop significantly, which verifies the usefulness of fusing information from different frames in TA. When compared with "B+M", "+CAM" achieves better results, which shows its capacity of fusing multi-modal features. Furthermore, we try to remove all CMA modules in TA and only adopt a concatenation with a convolutional layer to fuse multi-modal features before the MMVT, which is denoted as "+CAT+TA". This results in worse performance than "B+M+T", which verify that it is useful and necessary to combine CMA and TA in our MMVT. % ️️✔️

% \vspace{-1mm}
\noindent\textbf{Decoder Settings.} 
In Table~\ref{decoder abl}, we conduct experiments to explore the effectiveness of our LGFF. We adopt the concatenation operation followed by a  convolution layer to fuse appearance, motion, linguistic features as well as features from higher level as the baseline, which is denoted as "CAT". We find that such a simple fusion strategy degrades the performance obviously, compared with our model "LGFF". This means that it is necessary to design the LGFF to effectively fuse multi-modal features from different levels.% ️️✔️

% \vspace{-1mm}
\noindent\textbf{Effectiveness of Multi-modal Alignment Loss} 
We conduct experiments to verify the effectiveness of our multi-modal alignment loss. "l2am" denotes adopting $L_{\mathcal{AL}}$ and $L_{\mathcal{ML}}$ to align linguistic features with appearance and motion features, while "a2m" represents employing $L_{\mathcal{AM}}$ to align appearance and motion features. We also try to add two traditional binary cross-entropy losses to "B+M+T+L" for appearance and motion branch, respectively, which is denoted as "+bce". From Table~\ref{alignment ablation}, we can find that "+bce" can not bring obvious improvement to the performance. When we add "l2am" and "a2m" to "B+M+T+L", the performance is improved gradually, which verifies the effectiveness of "l2am" and "a2m".% ️️✔️

\begin{table}[t!]
    \centering
    \footnotesize
    \caption{Comparison of using different MMVT setting.}
    \vspace{-3mm}
	\begin{tabular}{cc|cc|c|cc|} 
	\hline
	\multicolumn{2}{|c|}{\multirow{2}{*}{Name}} & \multicolumn{2}{|c|}{Setting} & \multicolumn{1}{c|}{\textbf{mAP}}  & \multicolumn{2}{c|}{\textbf{IoU}} \\ \cline{3-7}

    \multicolumn{2}{|c|}{}  &  \multicolumn{1}{c|}{CMA}  &  \multicolumn{1}{c|}{TA}   &  \multicolumn{1}{c|}{0.5:0.95}      & \multicolumn{1}{c|}{Overall}  & \multicolumn{1}{c|}{Mean}   \\  \cline{1-7}
    
    \multicolumn{2}{|c|}{B+M}  & \multicolumn{1}{c}{} & \multicolumn{1}{c}{}  & \multicolumn{1}{|c|}{36.3}   & \multicolumn{1}{|c|}{63.5}  & \multicolumn{1}{|c|}{49.5}  \\

    \multicolumn{2}{|c|}{+CMA}  & \multicolumn{1}{c}{\ding{51}} & \multicolumn{1}{c}{}  & \multicolumn{1}{|c|}{36.8}   & \multicolumn{1}{|c|}{64.1}  & \multicolumn{1}{|c|}{50.6}  \\  
    
    \multicolumn{2}{|c|}{+CAT+TA}  & \multicolumn{1}{c}{} & \multicolumn{1}{c}{\ding{51}}  & \multicolumn{1}{|c|}{38.6}   & \multicolumn{1}{|c|}{64.8}  & \multicolumn{1}{|c|}{53.2}  \\  
    \multicolumn{2}{|c|}{B+M+T}  & \multicolumn{1}{c}{\ding{51}} & \multicolumn{1}{c}{\ding{51}} & \multicolumn{1}{|c|}{39.2}   & \multicolumn{1}{|c|}{64.8}  & \multicolumn{1}{|c|}{53.6} \\  \hline

    \end{tabular}
    \label{transformer ablation}
    \vspace{-0.2cm}

\end{table}

\begin{table}[t!]
    \centering
    \footnotesize
    \vspace{-2mm}
    \caption{Comparison of using different decoder settings.}
    \vspace{-3mm}
	\begin{tabular}{cc|cc|c|cc|} 
	\hline
	\multicolumn{2}{|c|}{\multirow{2}{*}{Name}} & \multicolumn{2}{|c|}{Setting} & \multicolumn{1}{c|}{\textbf{mAP}}  & \multicolumn{2}{c|}{\textbf{IoU}} \\ \cline{3-7}
	   
    \multicolumn{2}{|c|}{}  &  \multicolumn{1}{c|}{CAT}   &  \multicolumn{1}{c|}{LGFF}    &  \multicolumn{1}{c|}{0.5:0.95}      & \multicolumn{1}{c|}{Overall}  & \multicolumn{1}{c|}{Mean}   \\  \cline{1-7}
    
    \multicolumn{2}{|c|}{CAT}  & \multicolumn{1}{c}{\ding{51}} & \multicolumn{1}{c}{}  & \multicolumn{1}{|c|}{37.6}   & \multicolumn{1}{|c|}{63.5}  & \multicolumn{1}{|c|}{51.6}  \\  
    
    \multicolumn{2}{|c|}{LGFF}  & \multicolumn{1}{c}{} & \multicolumn{1}{c}{\ding{51}} & \multicolumn{1}{|c|}{41.1}   & \multicolumn{1}{|c|}{66.8}  & \multicolumn{1}{|c|}{54.8}  \\  \hline
    
    \end{tabular}
    \label{decoder abl}
    \vspace{-0.6cm}

\end{table}

\begin{table}[t!]
    \centering
    \footnotesize
    \caption{Comparison of using different Multi-modal Alignment Loss settings.}
    \vspace{-3mm}
	\begin{tabular}{cc|cc|c|cc|} 
	\hline
	\multicolumn{2}{|c|}{\multirow{2}{*}{Name}} & \multicolumn{2}{|c|}{Setting} & \multicolumn{1}{c|}{\textbf{mAP}}  & \multicolumn{2}{c|}{\textbf{IoU}} \\ \cline{3-7}
    \multicolumn{2}{|c|}{}  &  \multicolumn{1}{c|}{l2am}   &  \multicolumn{1}{c|}{a2m}    &  \multicolumn{1}{c|}{0.5:0.95}      & \multicolumn{1}{c|}{Overall}  & \multicolumn{1}{c|}{Mean}   \\  \cline{1-7}
    
    \multicolumn{2}{|c|}{B+M+T+L}  & \multicolumn{1}{c}{} & \multicolumn{1}{c}{} & \multicolumn{1}{|c|}{41.1}   & \multicolumn{1}{|c|}{66.8}  & \multicolumn{1}{|c|}{54.8}  \\  \hline
    \multicolumn{2}{|c|}{+bce}  & \multicolumn{1}{c}{} & \multicolumn{1}{c}{} & \multicolumn{1}{|c|}{41.2}   & \multicolumn{1}{|c|}{66.3}  & \multicolumn{1}{|c|}{54.8}  \\  \hline

    \multicolumn{2}{|c|}{+l2am}  & \multicolumn{1}{c}{\ding{51}} & \multicolumn{1}{c}{}  & \multicolumn{1}{|c|}{41.2}   & \multicolumn{1}{|c|}{67.1}  & \multicolumn{1}{|c|}{55.2}  \\  
    
    \multicolumn{2}{|c|}{B+M+T+L+A}  & \multicolumn{1}{c}{\ding{51}} & \multicolumn{1}{c}{\ding{51}} & \multicolumn{1}{|c|}{41.9}   & \multicolumn{1}{|c|}{67.3}  & \multicolumn{1}{|c|}{55.8}  \\  \hline

    \end{tabular}
    \label{alignment ablation}
    \vspace{-0.5cm}

\end{table}

\vspace{-2mm}
\subsection{Qualitative Results Comparison}
\vspace{-2mm} 
We visualize some representative samples generated from "B", "B+M", and "B+M+T+L+A" in Figure~\ref{abl_figure}. In some complex scenes like Figure~\ref{abl_figure} (a) and (b), there are multiple objects moving, leading to unsatisfying segmentation results from "B+M", which simply adopts concatenation to fuse multi-modal features. From Figure~\ref{abl_figure} (c), we can find that, when the motion information is adopted, although the model "B+M" can find the car, it still misclassifies some pixels from other cars as foreground, while "B+M+T+L+A" can generate more accurate mask. These examples show that our method can well incorporate and fuse appearance, motion and linguistic features together to locate the target object and generate more accurate masks. Figure~\ref{abl_figure} (d) demonstrates that our model can still accurately segment the target object without motion.

\vspace{-2mm}  
\section{Conclusion}
\vspace{-2mm}
In this paper, we propose a method to fuse and align multi-modal features for text-based video segmentation. First, we introduce the explicit motion information from optical flow maps to incorporate with appearance and linguistic features. Then, we design the MMVT to fuse multi-modal features between frames. Furthermore, we propose the LGFF module to progressively fuse multi-modal features from different feature levels. Finally, the multi-modal alignment loss is adopted to explicitly align multi-modal features to reduce the semantic gap between them.  Extensive experiments verify the effectiveness of each component in our method and demonstrate that our method can significantly outperform state-of-the-art methods on two popular datasets. % ️️

\vspace{-6mm}  
\paragraph{Acknowledgments:}
We thank Google TFRC for supporting us to get access to the Cloud TPUs. We thank CSCS (Swiss National Supercomputing Centre) for supporting us to get access to the Piz Daint supercomputer. We thank TACC (Texas Advanced Computing Center) for supporting us to get access to the Longhorn supercomputer and the Frontera supercomputer. We thank LuxProvide (Luxembourg national supercomputer HPC organization) for supporting us to get access to the MeluXina supercomputer. This research is supported by the National Research Foundation Singapore under its AI Singapore Programme (AISG Award No: AISG2-PhD-2021-08-008),
NRF Centre for Advanced Robotics Technology Innovation (CARTIN) Project,
and NUS Faculty Research Committee Grant (WBS: A-0009440-00-00).

{\small
\bibliographystyle{ieee_fullname}
\bibliography{egbib}
}

\end{document}